\documentclass[10pt,twocolumn,letterpaper]{article}
 
\usepackage{iccv}
\usepackage{times}
\usepackage{epsfig}
\usepackage{graphicx}
\usepackage{amsmath}
\usepackage{amssymb}
\usepackage{subfig}
\usepackage{siunitx}
\usepackage{pifont}
\usepackage{comment}
\usepackage{booktabs}
\usepackage[pagebackref=true,breaklinks=true,letterpaper=true,colorlinks,bookmarks=false]{hyperref}
\usepackage{enumitem}

\usepackage{authblk}

\usepackage[accsupp]{axessibility}  

\makeatletter
\renewcommand\AB@affilsepx{ \protect\Affilfont}
\makeatother

\iccvfinalcopy 

\def\iccvPaperID{4204} 

\ificcvfinal\fi

\def\etal{\textit{et al}.}
\def\ie{\textit{i.e.}\xspace}
\def\eg{\textit{e.g.}\xspace}
\newcommand{\name}{Shuffled-DBN\xspace}
\newtheorem{myDef}{Definition} 

\newcommand{\trieq}[0]{\triangleq}
\DeclareDocumentCommand{\expectunder}{O{}mO{}}{%
    \underset{{#1}}{\mathbb{E}}%
    \ifthenelse{\isempty{#2}}{}{\hspace{-3pt}\left[{#2}\ifthenelse{\isempty{#3}}{}{\mathrel{}\middle\vert\mathrel{}#3}\right]}
}
\newcommand*{\iidsim}{\overset{\text{i.i.d.}}{\sim}}
\newcommand*{\pdata}{p_\mathsf{neg}}
\newcommand*{\distndata}{\pdata}
\newcommand*{\ppos}{p_\mathsf{pos}}
\newcommand*{\distnpos}{\ppos}
\newcommand*{\lunif}{\mathcal{L}_\mathsf{uniform}}
\newcommand*{\lalign}{\mathcal{L}_\mathsf{align}}

\begin{document}

\title{On Feature Decorrelation in Self-Supervised Learning}

\author[1,2]{Tianyu Hua\thanks{Equal contribution.}}
\newcommand\CoAuthorMark{\footnotemark[\arabic{footnote}]}
\author[1]{Wenxiao Wang\protect\CoAuthorMark}
\author[2,3]{Zihui Xue}
\author[2,5]{Sucheng Ren}
\author[4]{Yue Wang}

\author[1,2]{Hang Zhao\thanks{Corresponding to hangzhao@mail.tsinghua.edu.cn.}}

\affil[1]{Tsinghua University~~}
\affil[2]{Shanghai Qi Zhi Institute\authorcr}
\affil[3]{UT Austin~~}
\affil[4]{MIT~~}
\affil[5]{South China University of Technology}

\date{}

\maketitle
\ificcvfinal\thispagestyle{empty}\fi

\begin{abstract}

In self-supervised representation learning, a common idea behind most of the state-of-the-art approaches is to enforce the robustness of the representations to predefined augmentations. 
A potential issue of this idea is the existence of completely collapsed solutions (\ie, constant features), which are typically avoided implicitly by carefully chosen implementation details. 
In this work, we study a relatively concise framework containing the most common components from recent approaches. 
We verify the existence of \textbf{complete collapse} and discover another reachable collapse pattern that is usually overlooked, namely \textbf{dimensional collapse}. 
We connect dimensional collapse with strong correlations between axes and consider such connection as a strong motivation for \textbf{feature decorrelation} (\ie, standardizing the covariance matrix). 
The 
gains from feature decorrelation are verified empirically to highlight the importance and the potential of this insight.


\end{abstract}


\section{Introduction}

\begin{figure}[!t]
\centering
\subfloat[complete collapse]{
\includegraphics[width=0.45\linewidth]{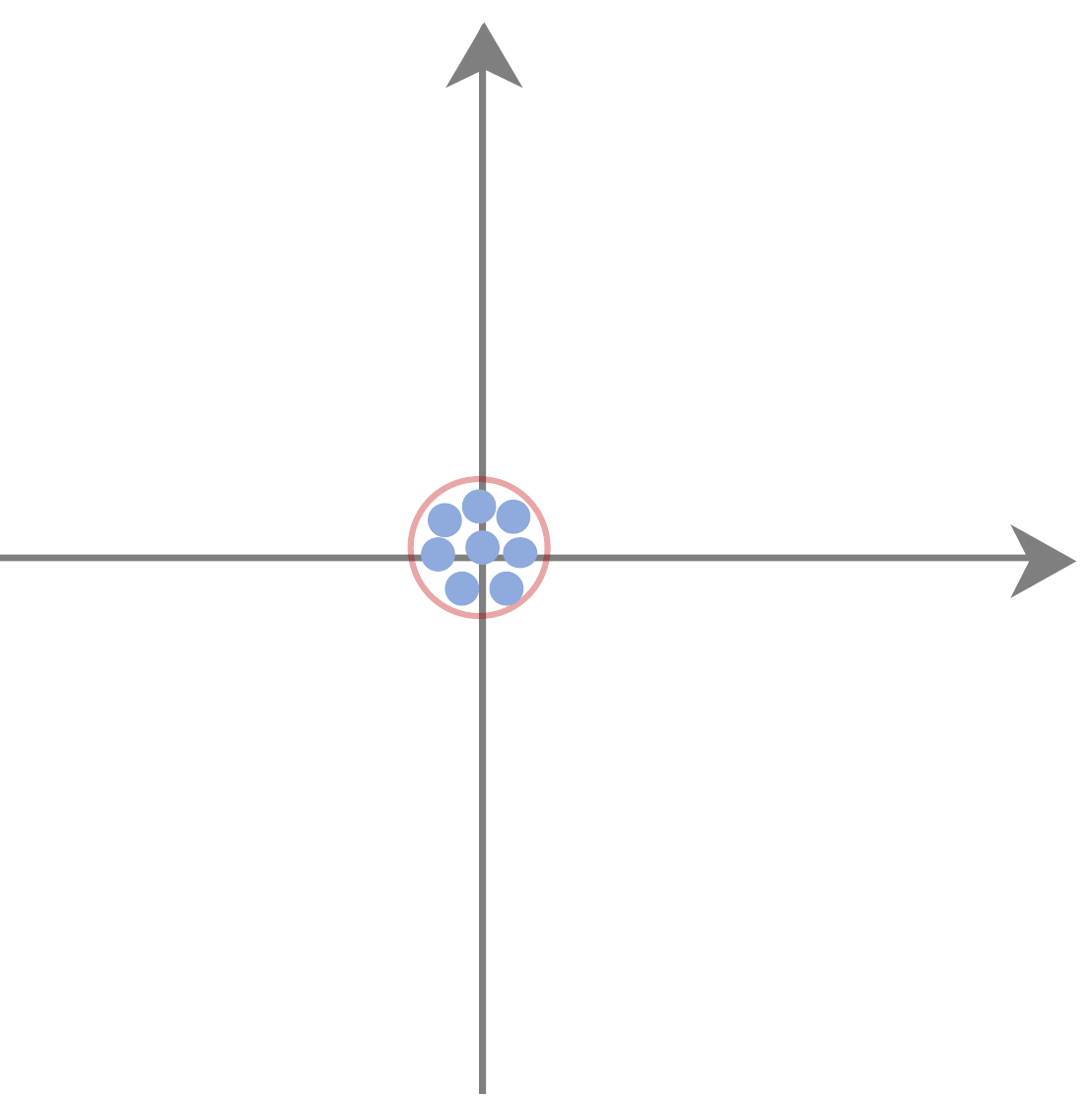}
\label{fig:col1}
}
\hfill
\subfloat[dimensional collapse]{
\includegraphics[width=0.45\linewidth]{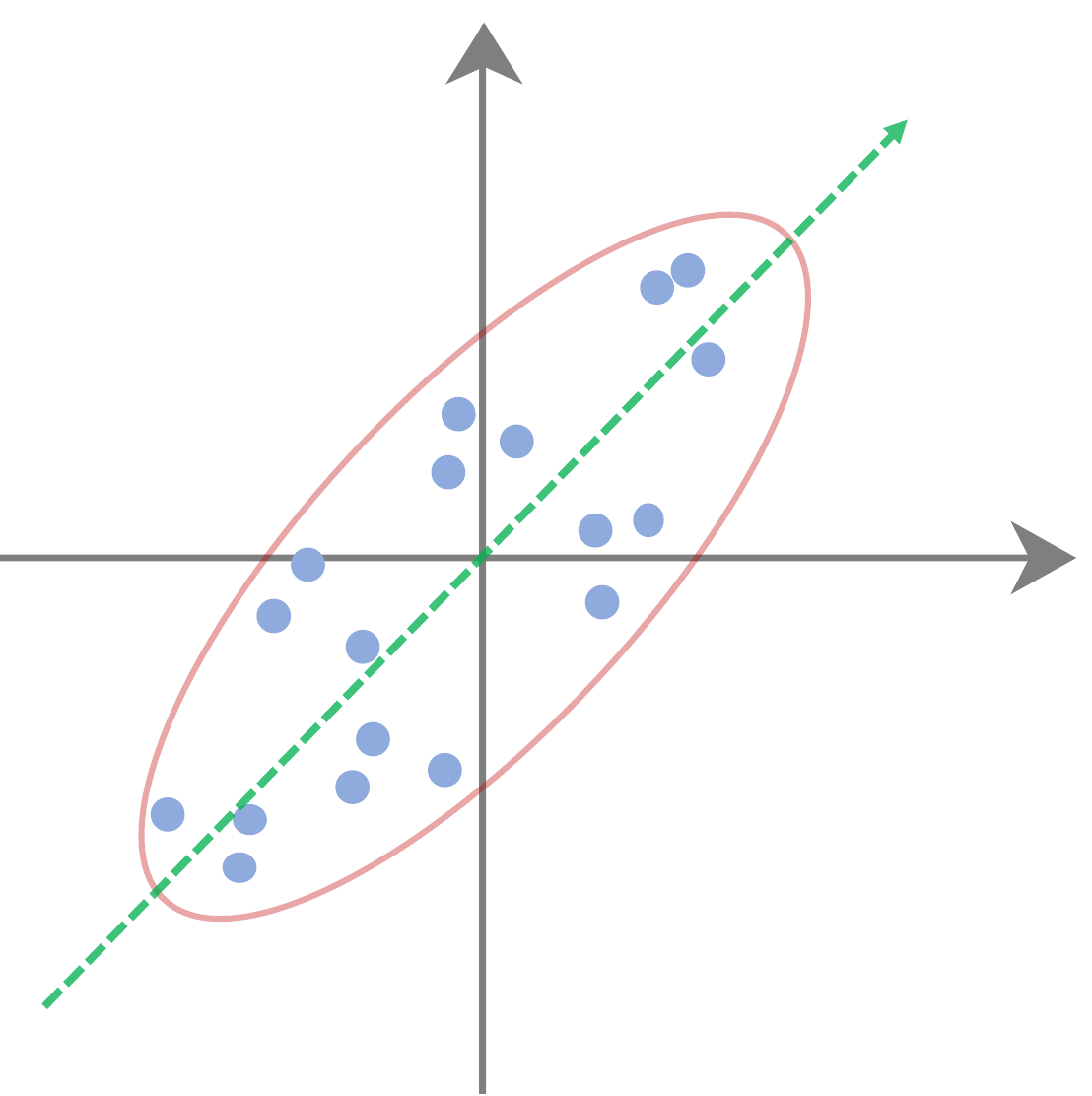}
\label{fig:col2}
}

\subfloat[decorrelated]{
\includegraphics[width=0.45\linewidth]{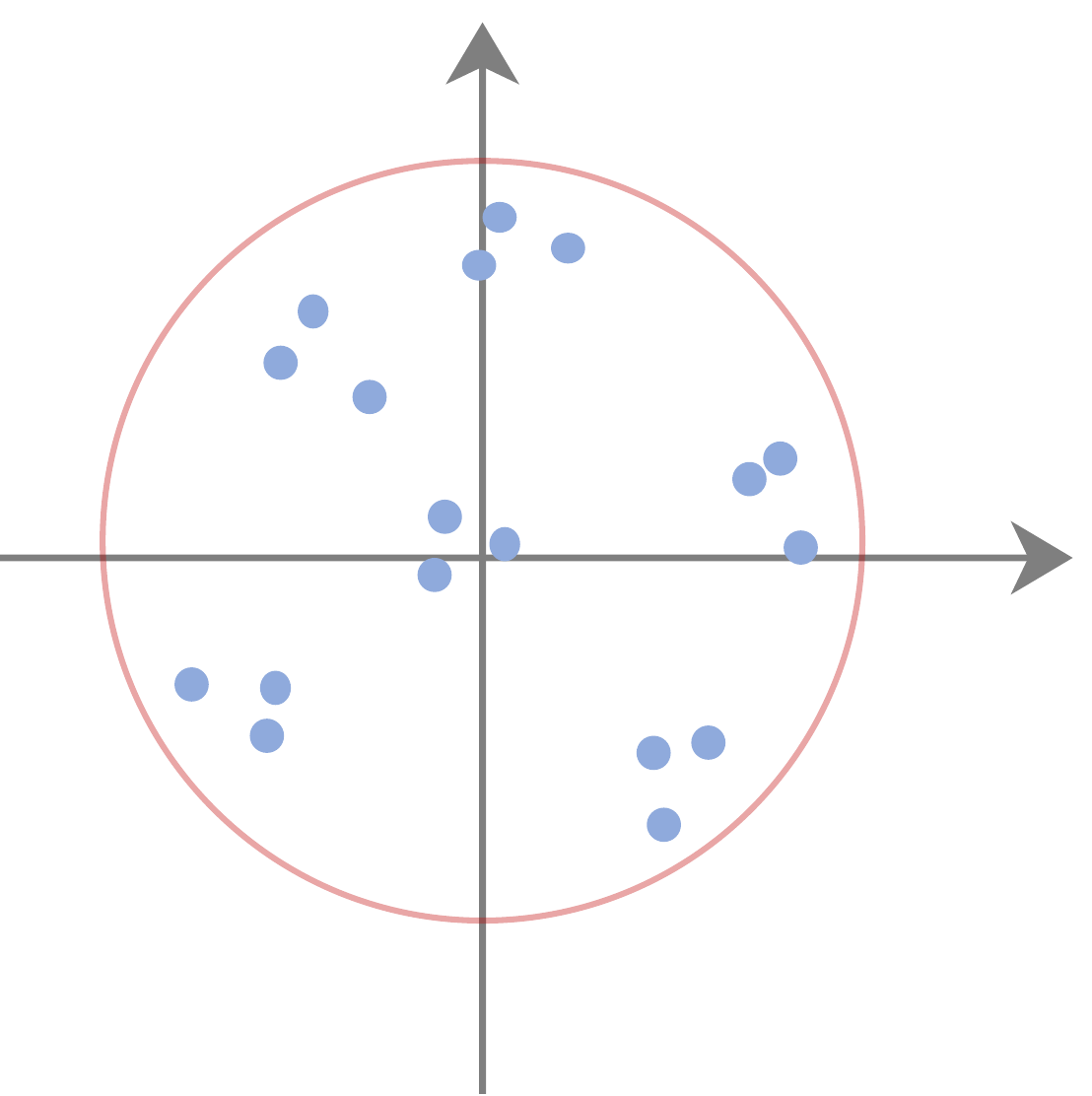}
\label{fig:col3}
}
\hfill
\subfloat[the concise framework]{
  \includegraphics[width=0.45\linewidth]{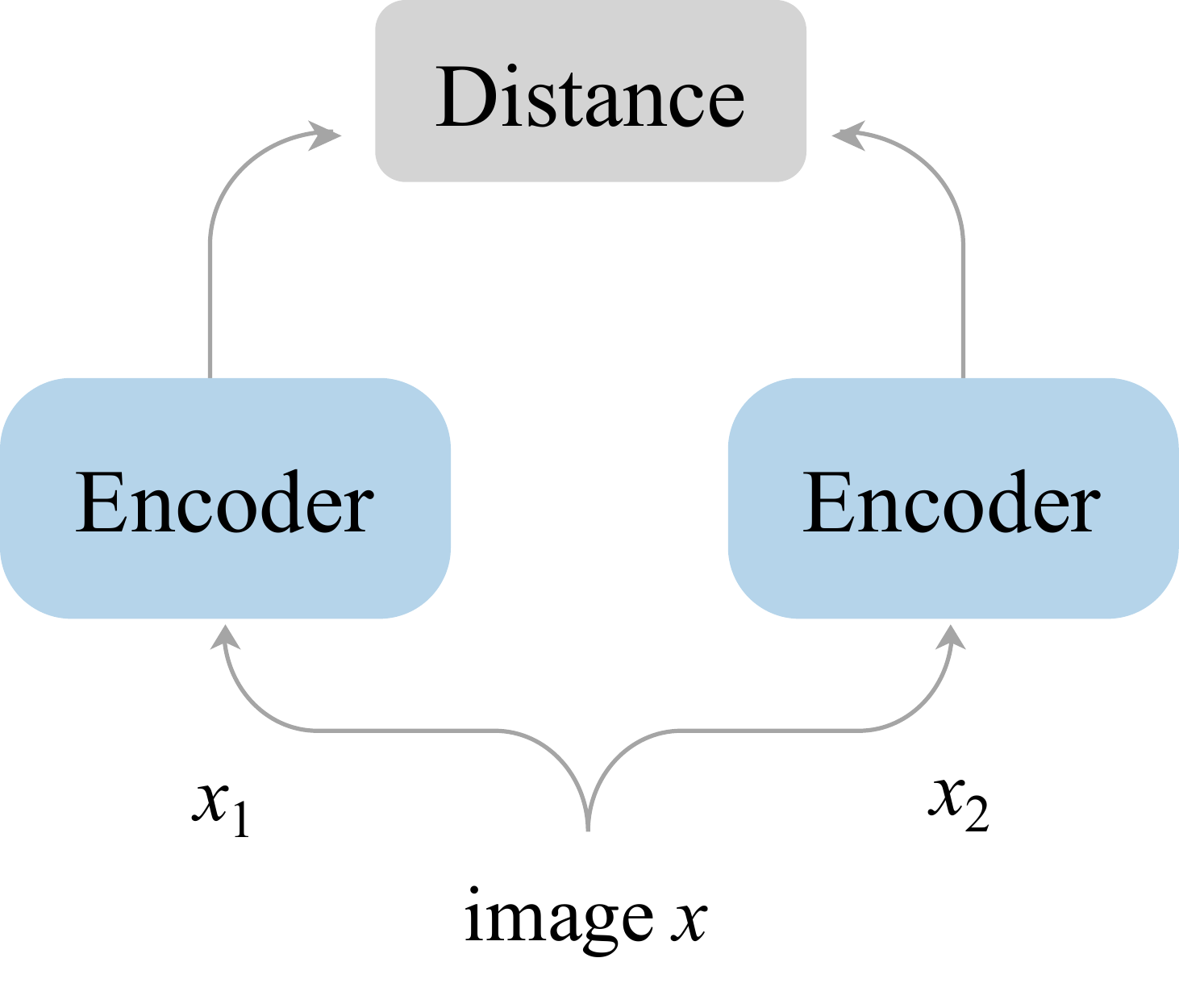}
  \label{fig:framework}
}

\caption{An overview of the key components of this work: \ref{fig:col1} and \ref{fig:col2} are two reachable collapse patterns in self-supervised settings; \ref{fig:col3} is an illustration of the goal of feature decorrelation; \ref{fig:framework} is a sketch of the concise framework used in this work.}
\label{fig:collapse}
\end{figure}

Deep learning is prevailing in a wide range of domains, including computer vision \cite{he2016deep}, natural language processing \cite{bert} and speech recognition \cite{SpeechRecog}, while the utility of the most classical, supervised methods are sometimes restricted by limited or costly data labeling. Recently, self-supervised learning has proven capable of offering visual representations with high utility and therefore reducing the need for massive annotations.
There has been significant advancements in this field in the past years: A line of work focuses on determining augmentations that better suit the self-supervised fashion, including revisiting typical augmentations \cite{tian2020makes}, using adversarial perturbations \cite{hu2020adco,SslHardNeg,ContrastAdv}, and searching augmentation policies \cite{selfaugment}; A line of work alters the sampling strategy to expand the source of positive pairs \cite{SeedTheViews,MYOW,InvarPropagate} and calibrate the contribution of negative samples \cite{InvarPropagate, ContrastHardNeg, FalseNegCancel}; 
Another line of work uses clustering-based mechanisms to characterize the relation of cross-sample views \cite{caron2018deepcluster, caron2019unsupervised, asano2019self, Caron2020swav}.
Despite the technical variety, there is one high-level idea that remains in most if not all of the recent approaches, learning representations that are robust to augmentations \cite{oord2018representation,chen2020simple,he2020moco,tian2020makes}.  

This idea is fairly intuitive but it does not rule out trivial, collapsed solutions by design. 
In consequence, existing work must incorporate a way that helps to mitigate the issue of feature collapsing.
Complete collapse, where the representations collapse into a constant as in Figure \ref{fig:col1}, is the most well-known type of collapse and is addressed differently in existing work with carefully chosen implementation details: To name a few, SimCLR \cite{chen2020simple} and Uniformity\cite{uniformity} use losses that maximize the distance between different samples; SwAV \cite{Caron2020swav} includes an additional online clustering branch that clusters data into a predefined number of groups; BYOL \cite{grill2020bootstrap} relies on a predictor structure, a properly inserted stop-gradient operator and a momentum encoder;  SimSiam \cite{chen2020exploring} simplifies the framework of BYOL by removing the momentum encoder. Given their success in preventing complete collapse, the study of other potential collapse issues in self-supervised learning has been ignored. 

Meanwhile, feature decorrelation appears to be a valuable idea in the field of machine learning: In discriminative tasks,  \cite{ReduceOverfitDecor,StructuredDecor} introduce in objective functions additional terms regularizing correlation matrices  and \cite{huang2018decorrelated, huang2019iterative} develop normalization layers standardizing covariance matrices to obtain higher accuracy;
In generative tasks, it is through feature decorrelation that \cite{WhitenColorGAN} produces more realistic synthesized images and \cite{UDA_Whiten} offers better utility of domain adaptation.

In this work, we revisit the collapse issue of self-supervised learning and show how the idea of feature decorrelation helps to resolve the issue and improve the utility, using a framework presented in Figure \ref{fig:framework} that contains the most common components of existing approaches. 
Our contribution includes:
\begin{itemize}[wide=0pt]
\setlength\itemsep{-0.25em}
\item We verify the existence of complete collapse in self-supervised settings and address it successfully by standardizing variance. 
\item We discover another reachable collapse pattern ignored by existing works, namely dimensional collapse.
\item We reveal the connection between dimensional collapse and strong correlations, which leads to the idea of standardizing covariance (\ie feature decorrelation).
\item Empirically, the performance gains from feature decorrelation in a wide range of settings confirm the importance and the potential of this insight.
\end{itemize}

\section{Related Work}

\textbf{Contrastive learning.} Contrastive approaches learn representations by maximizing agreement between two augmented views of a sample (\ie, positive pairs) and disagreement of views from different samples (\ie negative pairs). Following this idea, many methods have been developed \cite{oord2018representation, hjelm2018learning, wu2018unsupervised, henaff2020data, he2020moco, chen2020simple, uniformity}. As they benefit from a large number of negative samples, contrastive learning methods require a memory bank \cite{wu2018unsupervised}, a queue \cite{he2020moco} to store negative samples, or large batch sizes \cite{chen2020simple} to work well. This leads to the question of whether using negative samples is necessary.

\textbf{Clustering.} Clustering-based methods partially answer this question. They discriminate between groups of images with similar features instead of individual images \cite{caron2018deepcluster, caron2019unsupervised, asano2019self, Caron2020swav}. SwAV \cite{Caron2020swav} clusters data and enforces consistency between cluster assignments produced from different views of the same sample. However, these methods require a costly clustering phase and large batches to have a sufficient number of samples for clustering \cite{grill2020bootstrap, chen2020exploring}.

\textbf{BYOL and SimSiam.} Another recent line of work achieves remarkable results by only using positive samples. BYOL \cite{richemond2020byol} proposes an online network along with a target network, where the target network is updated with a moving average of the online network to avoid collapse. Contrary to them, SimSiam \cite{chen2020exploring} demonstrates that a predictor network and a properly inserted stop-gradient operator are the crucial components in preventing collapse. Tian \etal \ provide an analysis of how various factors involved in BYOL and SimSiam work together to prevent collapse \cite{tian2021understanding}.  

\textbf{Normalization.} Different from previous works that attribute collapse prevention to ``asymmetry'', \ie, a predictor network and stop gradient, we propose a new angle to understand collapse in this work. Based on this view, we introduce normalization techniques in supervised learning to the task of learning representations without negative pairs. Batch Normalization (BN) \cite{ioffe2015batch} is the first to perform normalization per mini-batch in a way that supports back-propagation, and has shown remarkable performance in training deep neural networks. The idea of BN is to center and scale activations. Another normalization technique, Decorrelated Batch Normalization (DBN) \cite{huang2018decorrelated} proposes to whiten activations within each mini-batch. In this work, we demonstrate that in the context of self-supervised learning, BN encounters 
dimensional collapse while DBN effectively avoids all kinds of collapse. 

Two concurrent works \cite{ermolov2020whitening,zbontar2021barlow} explore similar ideas to ours for preventing collapses. W-MSE \cite{ermolov2020whitening} whitens feature representations within each batch via Cholesky decomposition. Barlo Twins \cite{zbontar2021barlow} enforces the cross-correlation matrix between outputs of a positive pair to be close to identity, using an additional loss function. These attempts corroborate the potential of feature decorrelation and highlight the necessity of our findings towards understanding and addressing feature collapses.


\section{Main Results}
\begin{figure*}[!tbp]
    \centering
    \subfloat[baseline: complete collapse]{
        \includegraphics[width=0.18\linewidth]{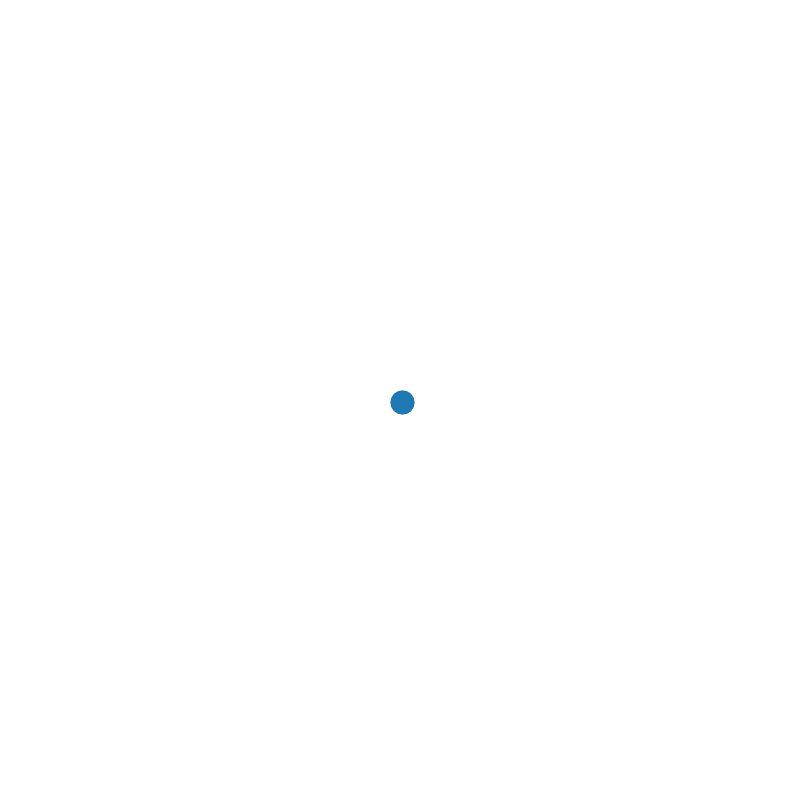}
        \label{fig:vis_linear}
    }
    \hfill
    \subfloat[Batch Normalization: dimensional collapse]{
        \includegraphics[width=0.18\linewidth]{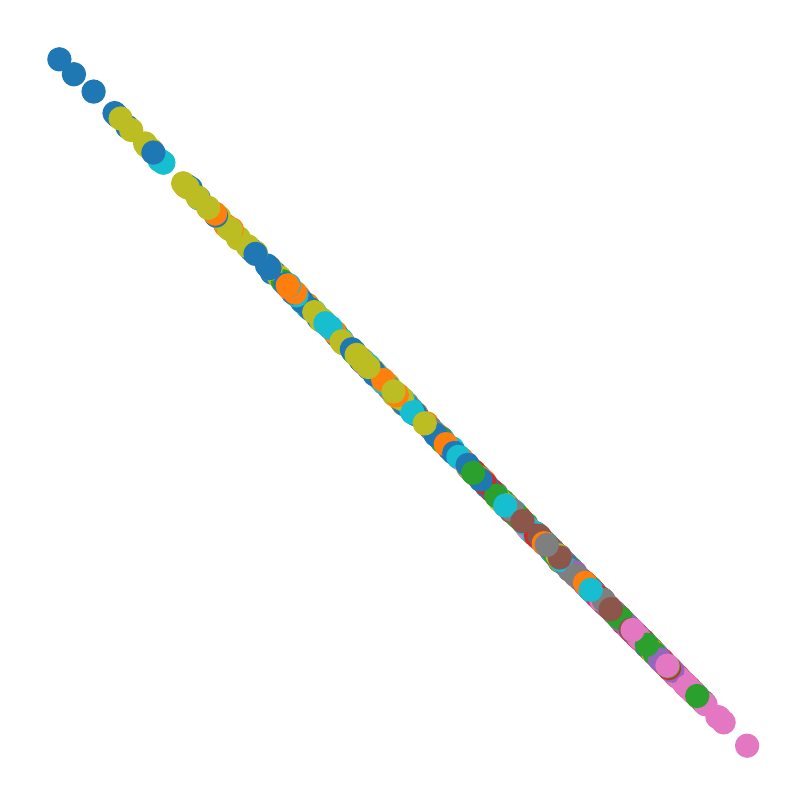}
        \label{fig:vis_BN}
    }
    \hfill
    \subfloat[Decorrelated BN: decorrelated space]{
        \includegraphics[width=0.18\linewidth]{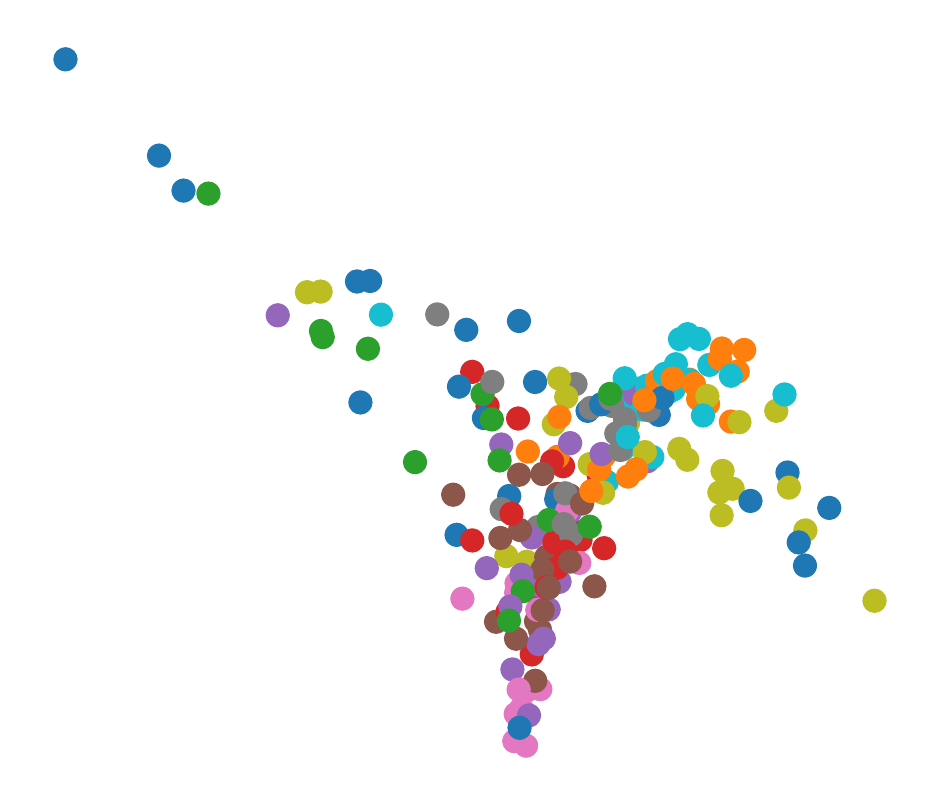}
        \label{fig:vis_DBN}
    }
    \hfill
    \subfloat[SimCLR]{
        \includegraphics[width=0.18\linewidth]{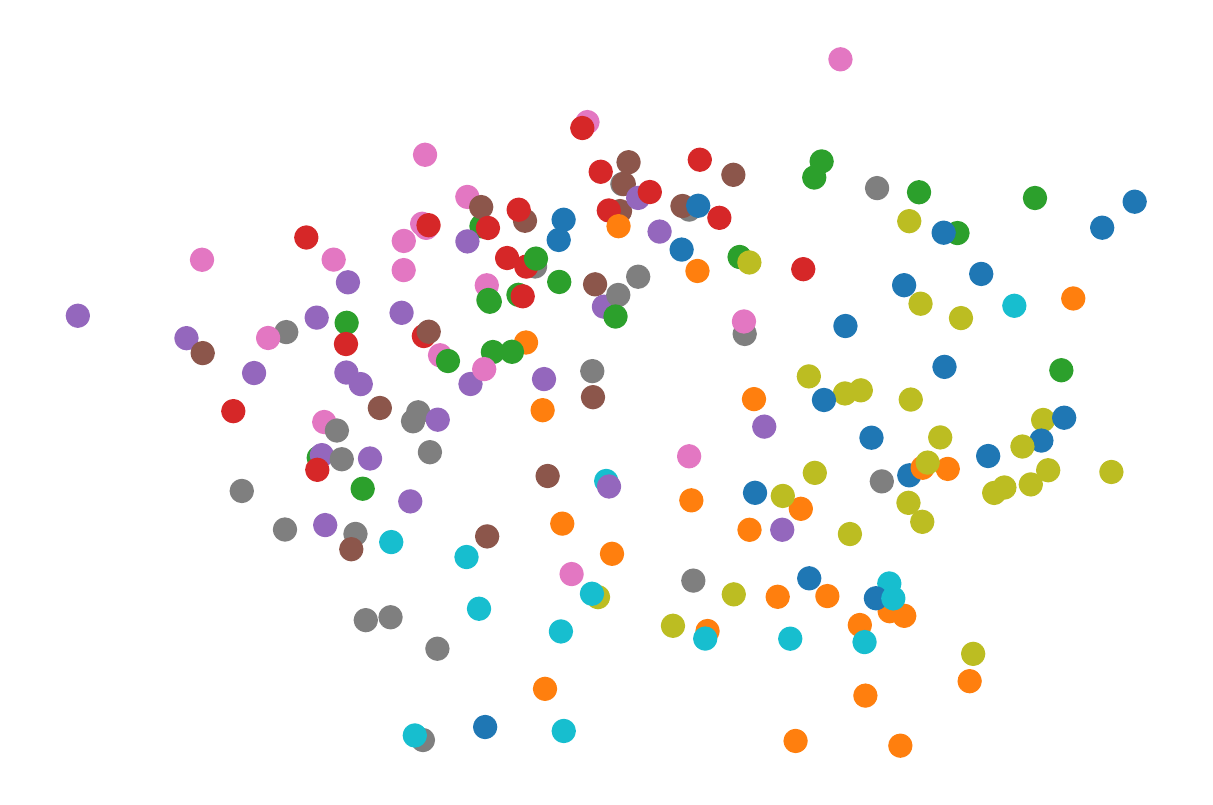}
        \label{fig:vis_simclr}
    }
    \hfill
    \subfloat[Supervised]{
        \includegraphics[width=0.18\linewidth]{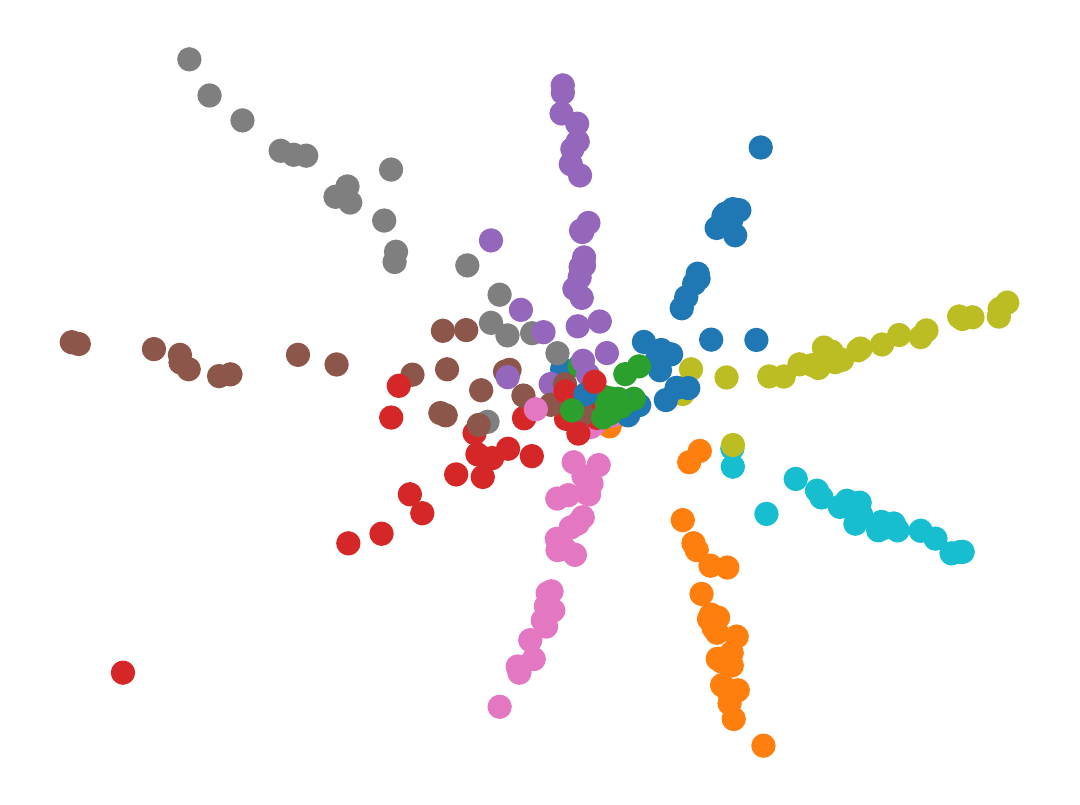}
        \label{fig:vis_supervised}
    }
    
   \caption{
   Direct visualization of $2$-dimensional projection spaces on CIFAR-10. Different colors correspond to different classes. 
   Figure \ref{fig:vis_linear}, \ref{fig:vis_BN} and \ref{fig:vis_DBN} are from the concise framework we use.
   For completeness, we visualize the $2$-dimensional projection spaces of SimCLR (by setting the output dimension of the projector to be $2$) and a supervised baseline (by letting the penultimate layer to contain $2$ neurons) in Figure \ref{fig:vis_simclr} and \ref{fig:vis_supervised}.
   }
   \label{fig:viz_feature}
\end{figure*}

In this section, we will take a close look at two collapse patterns in self-supervised learning settings.
We will show how the first one, a well-known collapse pattern termed \textbf{complete collapse}, is addressable with BN \cite{ioffe2015batch} since it is associated with vanishing variances.

Furthermore, with complete collapse avoided, we discover another reachable collapse pattern overlooked by existing work, namely \textbf{dimensional collapse}. We relate dimensional collapse to strong correlations between axes and show with DBN \cite{huang2018decorrelated} that standardizing covariance matrix helps in alleviating dimensional collapse.

We also introduce in this section an easy add-on to DBN that enforces further decorrelation, which will be compared empirically with DBN in Section \ref{sec:experiment} to support the importance and the potential of feature decorrelation. We refer to DBN with this add-on as \name.

In the last part of this section, we include elaborations of some details that are postponed for coherence.

\subsection{Preliminary}
We utilize here a relatively concise framework for self-supervised representation learning as follows, which contains only the most common components of modern self-supervised approaches: 
\begin{myDef}[Concise Framework]

In the concise framework, given the training data distribution $\mathcal{D}$ and the augmentation distribution $\mathcal{T}$, the model parameter $\theta$ is trained to maximize/minimize the objective function with the following form:
\begin{align*}
    \mathcal{L}(\theta) = \mathbb{E}_{\substack{x\sim \mathcal{D} \\ T_1, T_2\sim \mathcal{T} } } \ell( f_\theta\left( x_1\right),~  f_\theta\left( x_2\right)),
\end{align*}
where $f_{\theta}$ is the encoder that contains a backbone and a projector, $x_1 = T_1(x)$, $x_2 = T_2(x)$, and $\ell$ is the similarity/distance function. A sketch of this framework is presented in Figure \ref{fig:framework}.
\label{def:framework}
\end{myDef}
Unless otherwise specified, squared error $\ell(z_1, z_2) = \| z_1 - z_2 \|^2_2$ is used as the distance function. We will elaborate the choice of $\ell$ in Section \ref{sec:loss}.

\subsection{Reachable Collapse Patterns and Their Indicators}

To build up intuitions, we now apply to CIFAR-10 a specific realization of the concise framework, which we refer to as the baseline: 
The encoder $f_\theta$ is a ResNet-18 backbone plus a projector MLP with an output dimension of 2 and two hidden layers with $64$ neurons in each. 
ReLU activation and BN are appended to both hidden layers of the projector.

Here we set the output dimension of the projector to be 2 for easy visualization. 
We will show later that our findings remain in high-dimensional projection space.

The resulted representation yields an accuracy of only $28.56\%$ in linear evaluation and we visualize in Figure \ref{fig:vis_linear} the projection space (\ie $f_\theta(T(x))$ where $x\sim \mathcal{D}, T\sim \mathcal{T}$).

In Figure \ref{fig:vis_linear}, we observe a projection space that collapses to a single point, which we refer to as complete collapse. When it happens, almost no gradient can be propagated back through the projection space (since $\nabla_{  f_\theta(T(x))} {\ell} \approx 0$) to influence the learned representation and therefore its utility is compromised.

Complete collapse is a widely known type of collapse in representation learning, and it is associated with \textbf{vanishing variances}. Accordingly, using BN in the projection space to standardize variance can be a way to mitigate complete collapse. 

\begin{myDef}[Batch Normalization \cite{ioffe2015batch}]
For a Batch Normalization (BN) layer that takes as its input a batch of $D$-dimensional vectors $X=\left(x_1, \cdots, x_B\right) \in \mathcal{R}^{D\times B} $, its output is a batch of vectors $Y=\left(y_1 \cdots, y_B\right) \in \mathcal{R}^{D\times B}$, computed as follows: 
\begin{align*}
    y_{b, d} = \frac{x_{b, d} - \mu_{d}}{\sqrt{\sigma^2_{d} + \epsilon}} \cdot \gamma_{d} + \beta_{d}
\end{align*}
for all $b\in \{1, \cdots, B\}$ and $d\in \{1, \cdots, D\}$, 
where $\gamma, \beta$ are learnable affine parameters, $\epsilon$ is a small constant originally proposed for numerical stability. 
In training time, $\mu_d, \sigma^2_d$ are mean and variance computed over the $d$-th row of the input batch $X$, and in inference time, running estimations from training time are used.
\end{myDef}

We append to the projector of the baseline an additional BN layer with no affine parameter and $\epsilon=0$ (\ie $ y_{i, j} = \frac{x_{i, j} - \mu_{j}}{\sqrt{\sigma^2_{j} }}$, whose necessity will be elaborated in Section \ref{sec:BN_affine_eps}) and visualize the projection space in Figure \ref{fig:vis_BN}. 
The corresponding representation yields an accuracy of $69.52\%$ in linear evaluation, significantly improved over $28.56\%$ obtained by the completely collapsed baseline.

With complete collapse resolved, we notice another usually overlooked collapse pattern in the projection space, termed dimensional collapse, for which the projected features collapse into a low-dimensional manifold such as the single line in Figure \ref{fig:vis_BN}. 
Dimensional collapse can harm utility and should be addressed appropriately. 
By definition, dimensional collapse is associated with \textbf{strong correlations} between axes. 
As a sanity check, we adapt DBN \cite{huang2018decorrelated} to standardize the covariance matrix for mitigation of this issue.

\begin{myDef}[Decorrelated Batch Normalization \cite{huang2018decorrelated}]
The Decorrelated Batch Normalization (DBN) layer with a group size $G$ takes as its input a batch of $D$-dimensional vectors $X=\left( x_1, \cdots, x_B\right) \in \mathcal{R}^{D\times B}$ and its output is a batch of vectors $Y=\left(y_1 \cdots, y_B\right) \in \mathcal{R}^{D\times B}$ computed as follows:
\begin{align*}
    Y^{[h]} = ZCA(X^{[h]}),
\end{align*}
where $X^{[h]} = \left( \left(X_{ (h-1) \cdot G + 1 } \right)^T, \cdots, \left(X_{h \cdot G} \right)^T \right)^T \in \mathcal{R}^{G \times B}$ and $Y^{[h]} = \left( \left(Y_{ (h-1) \cdot G + 1 } \right)^T, \cdots, \left(Y_{h \cdot G} \right)^T \right)^T \in \mathcal{R}^{G \times B}$. 
In other words, DBN divides the D feature dimensions into groups of size $G$ and applies ZCA whitening to each group independently.
\end{myDef}

\begin{myDef}[ZCA Whitening \cite{ZCA}]
ZCA Whitening takes as its input a batch of $D$-dimensional vectors $X=\left( x_1, \cdots, x_B\right) \in \mathcal{R}^{D\times B}$ and its output is a batch of vectors $Y=\left(y_1 \cdots, y_B\right) \in \mathcal{R}^{D\times B}$ computed as follows: 
\begin{align*}
    Y = Q\Lambda^{-\frac{1}{2}} Q^T\hat{X},
\end{align*}
where $\hat{X}$ is $X$ with rows normalized to zero-mean (\ie $\hat{X}_{d, b} = X_{d, b} - \frac{1}{B} \sum_{k=1}^B X_{d, k} = x_{b,d} - \frac{1}{B} \sum_{k=1}^B x_{k, d})$, $\Lambda \in \mathcal{R}^{D \times D}$ is a diagonal matrix filled with the eigenvalues of $\Sigma = \hat{X} \hat{X}^T$ and $Q\in\mathcal{R}^{D\times D}$ is the corresponding orthonormal eigenvectors (\ie $\Sigma = Q\Lambda Q^T$). 
ZCA assumes $\Sigma = \hat{X} \hat{X}^T \in \mathcal{R}^{D\times D}$ is full-rank.
\end{myDef}

The rows of the ZCA's output $Y$ are zero-mean, and therefore the corresponding covariance matrix is 
\begin{align*}
    YY^T = & Q\Lambda^{-\frac{1}{2}} Q^T\hat{X} \hat{X}^T Q\Lambda^{-\frac{1}{2}} Q^T \\
    =&Q\Lambda^{-\frac{1}{2}} Q^T Q \Lambda Q^T Q\Lambda^{-\frac{1}{2}} Q^T\\
    =& Q\Lambda^{-\frac{1}{2}}  \Lambda \Lambda^{-\frac{1}{2}} Q^T\\
    =&QQ^T = I.
\end{align*}
Thus DBN standardizes the covariance matrices of dimension groups to alleviate dimensional collapse issues.

We visualize in Figure \ref{fig:vis_DBN} the projection space with a DBN layer (group size $G=2$) appended to the projector of the baseline. The corresponding representation offers an accuracy of $72.45\%$ in linear evaluation, which reveals already, in this $2$-dimensional case, a non-negligible gap from $69.52\%$ offered by the dimensionally collapsed one. 
It corroborates the importance of feature decorrelation to self-supervised representation learning.

   \begin{figure}[t]
      \centering
      \subfloat{{\includegraphics[width=0.48\linewidth]{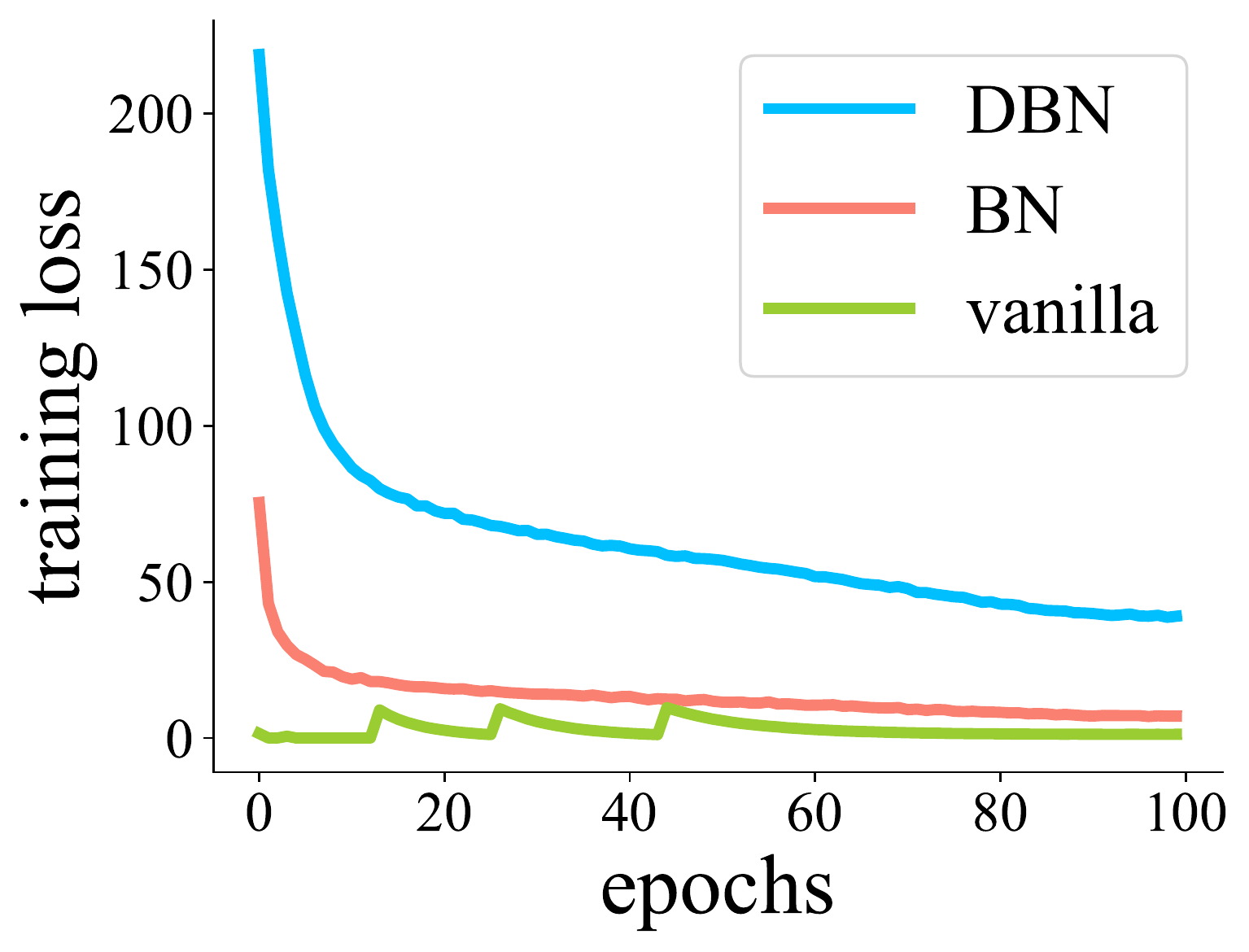}}}
      \hfill
      \subfloat{{\includegraphics[width=0.48\linewidth]{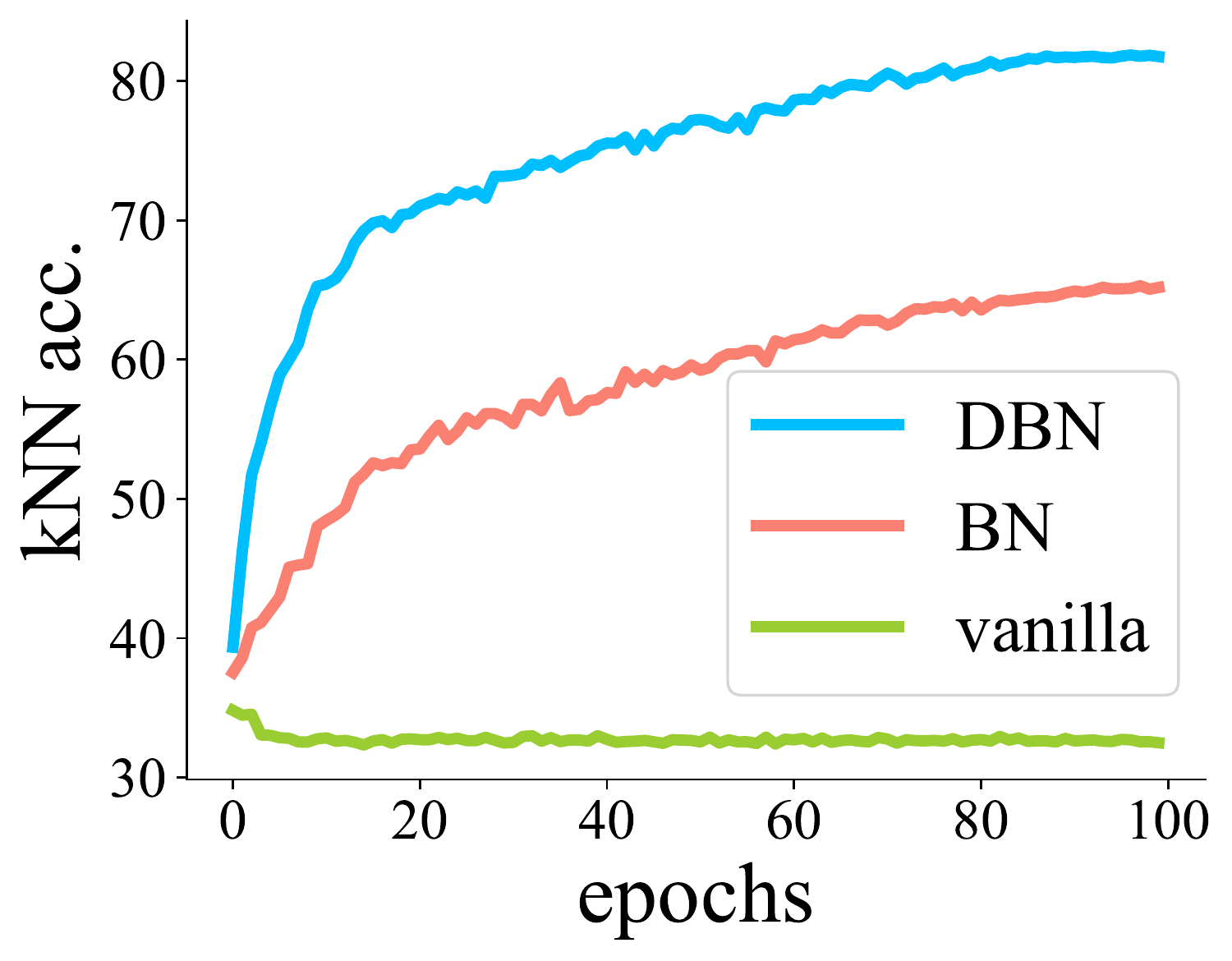}}}
      \caption{
      A comparison of learning processes with different variants of the concise framework. 
      For both collapse patterns, collapsed variants optimize loss function easily but offer representations with degraded utility.
      }
      \label{fig:linear_bn_dbn}

   \end{figure}
   
\begin{table}[!t]
    \centering
      \begin{tabular}{c c c c c}
      \toprule
             & acc. (\%) & std.    &  corr. & loss \\ 
      \midrule
      vanilla & 35.44  & \textbf{0.00} & 0.13           & 0.00\\
      BN     & 70.85 &  1.00       & \textbf{0.99}           & 7.01\\  
      DBN    & 84.41 &  1.00       & 0.00     & 39.04\\
      \bottomrule
      \end{tabular}
      \captionof{table}{
      A comparison of the eventual representations with different variants of the concise framework:
      \textbf{acc.} denotes the accuracy in linear evaluation; \textbf{std.} denotes the average standard deviation over $128$ dimensions of the projected features; \textbf{corr.} denotes the average correlation strength (\ie the average of the absolute values of non-diagonal entries of the correlation matrix) of the projected features; \textbf{loss} denotes the training loss. 
      The group size of DBN is  $128$.
      }
      \label{tab:linear_bn_dbn}
   \end{table}

The collapse patterns we observed remain reachable with a high-dimensional projection space and remain linked respectively with vanishing variances and strong correlations. 
In Figure \ref{fig:linear_bn_dbn} and Table \ref{tab:linear_bn_dbn}, we include a comparison of these variants when the projector is a $2$-layer MLP with $128$ hidden neurons and $128$-dimensional outputs.

In this comparison, we observe vanishing variances (through std.) with the vanilla framework and strong correlations (through corr.) with BN, which serve as signs of complete collapse and dimensional collapse, respectively. 
Another observation is that the utility gain by addressing these collapse patterns enlarges with an increased dimension of projection spaces. This observation further corroborates the potential of feature decorrelation.

\subsection{Further Decorrelation, Further Gains}
We show in the previous section that feature decorrelation (\ie standardizing covariance matrix) alleviates an overlooked pattern of collapse named dimensional collapse and therefore improves utility. 

However, the dimensional collapse issue partially remains since DBN introduces a grouping strategy (whose necessity is elaborated in Section \ref{sec:design_choices}) that standardizes only covariances within each dimension group. 
To reveal the potential of feature decorrelation, we propose a variant of DBN with one easy add-on, namely \name. 
In this section, we show that \name offers further decorrelation. 
A more thorough evaluation of the further gains from the further decorrelation is included in Section \ref{sec:experiment}.

\begin{myDef}[\name]
The \name layer with a group size $G$ takes as its input a batch of $D$-dimensional vectors $X=\left( x_1, \cdots, x_B\right) \in \mathcal{R}^{D\times B}$ and its output is a batch of vectors $Y=\left(y_1 \cdots, y_B\right) \in \mathcal{R}^{D\times B}$ computed as follows:
\begin{align*}
    Y = \mathcal{P}^{-1} \left( DBN_G( \mathcal{P}(X) ) \right),
\end{align*}
where $\mathcal{P}$ is a random $D$-order permutation,  $\mathcal{P}(X)$ is obtained by rearranging rows of $X$ according to $\mathcal{P}$ and $\mathcal{P}^{-1}(X)$ is obtained by rearranging rows according to the inverse permutation of $\mathcal{P}$. 

In other words, \name permutes the D feature dimensions randomly before applying DBN with the same group size $G$ and reverses the permutation for outputs.
\end{myDef}

\begin{figure}[!t]
   \centering
    \subfloat[\textbf{corr.} denotes the average correlation strength (\ie the average of the absolute values of non-diagonal entries of the correlation matrix) of the projected features.]{
        \includegraphics[width=0.45\linewidth]{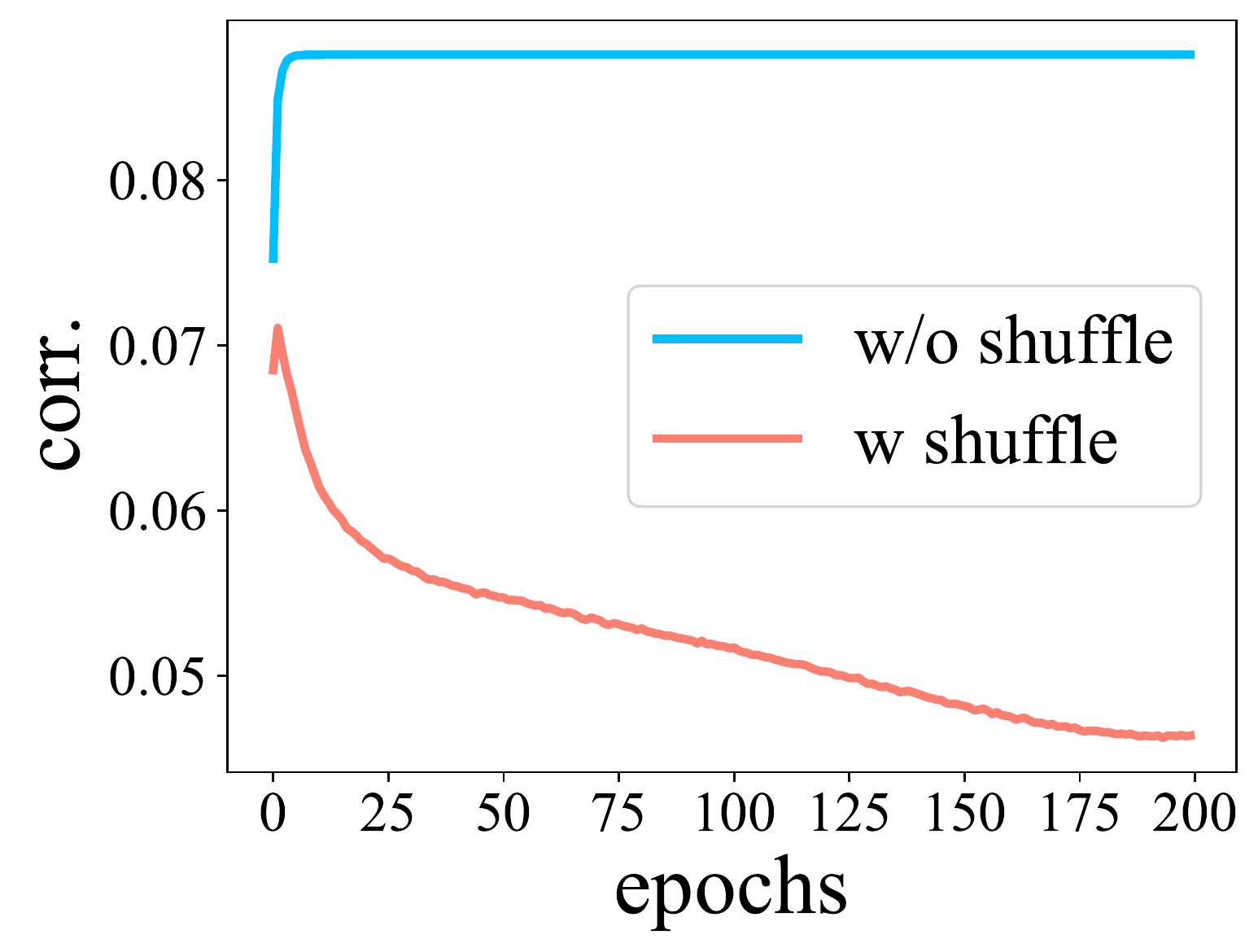}
    }
    \hfill
    \subfloat[ \textbf{rank} denotes the (estimated) rank of spaces spanned by projected features of 512 samples, which is computed by checking singular values.]{
        \includegraphics[width=0.45\linewidth]{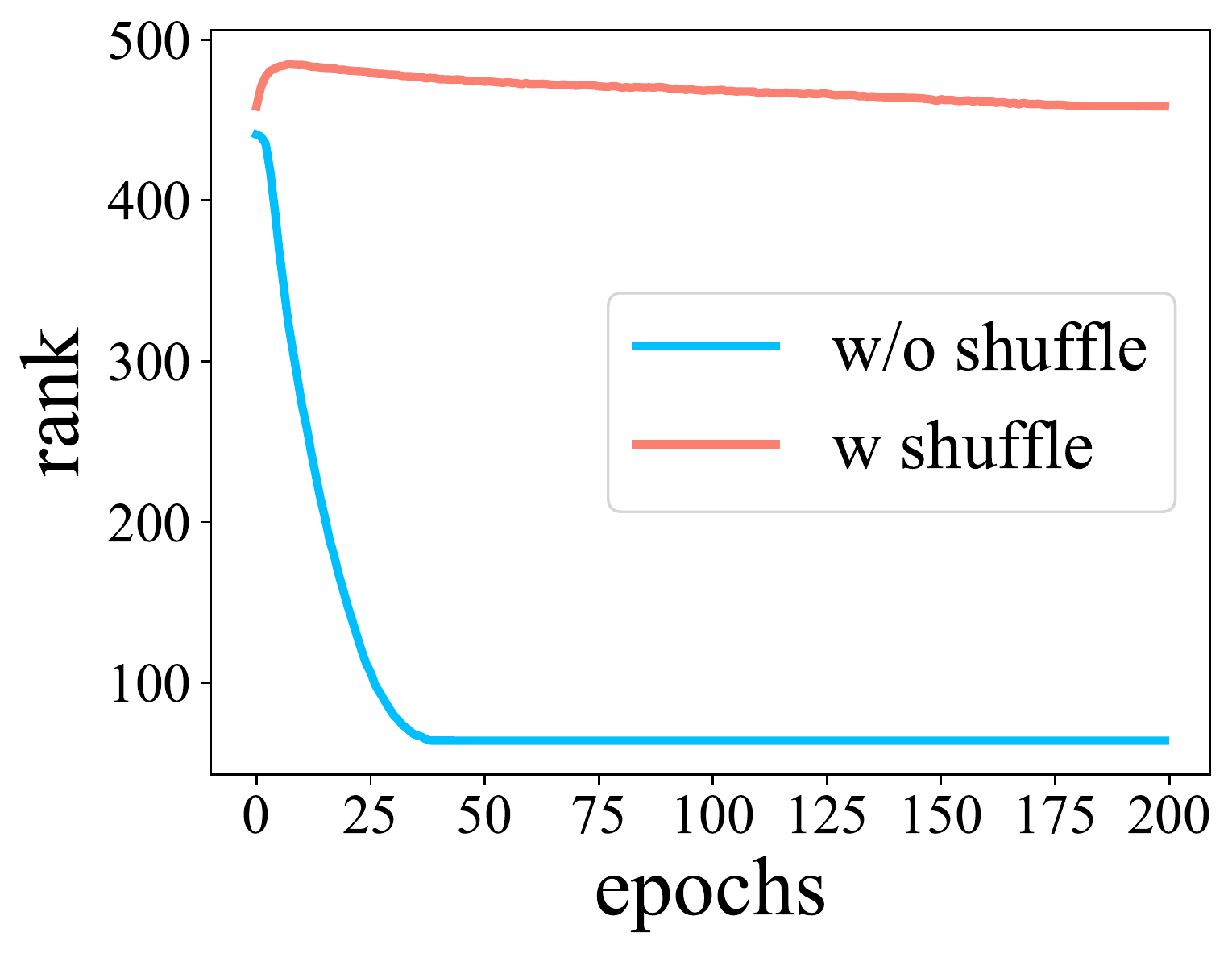}
    }
    
    \subfloat[\textbf{acc.} denotes accuracy in kNN classification.]{
        \includegraphics[width=0.45\linewidth]{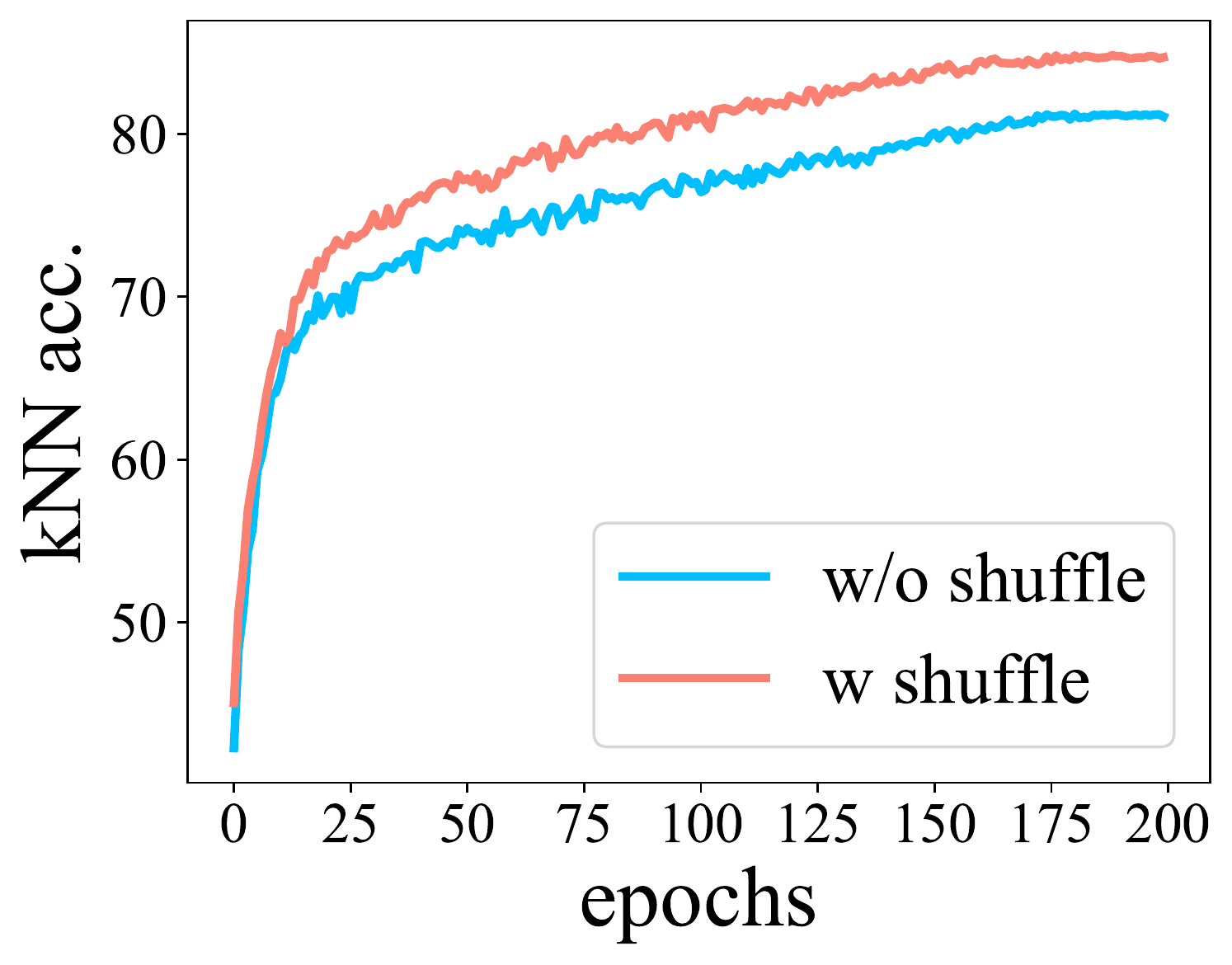}
    }
    \hfill
    \subfloat[ \textbf{loss} denotes the training loss.]{
        \includegraphics[width=0.45\linewidth]{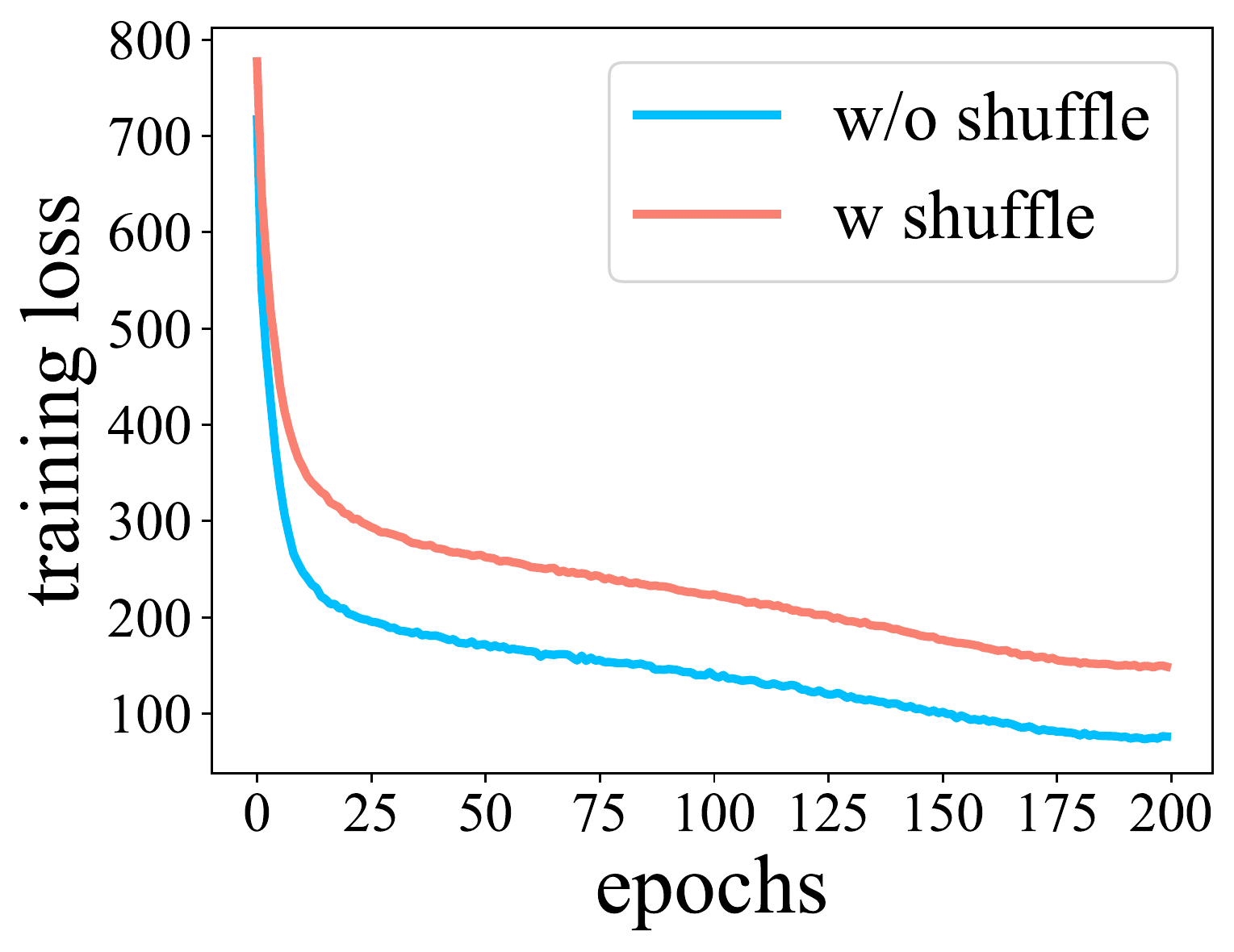}
    }

   \caption{A comparison of DBN (\ie w/o shuffle) and \name (\ie w shuffle). Compared with DBN, \name standardizes covariance matrix better (lower corr.), mitigates dimensional collapse more thoroughly (higher rank), and offers better utility (higher acc.). The group size is $64$ in both cases, and an MLP with two hidden layers (with respectively 512 and 1024 neurons) and an output dimension of 512 is used as the projector.}
   \label{fig:further_decorr}
\end{figure}

Intuitively, \name enforces further decorrelation since now each dimension is whitened with another $G-1$ randomly chosen dimensions rather than fixed ones, which standardizes the covariance matrix better. We verify the intuition empirically and include the results in Figure \ref{fig:further_decorr}, where \name offers lower correlation strength, less dimensional collapse, and better utility as expected. 
These support our claim regarding further decorrelation of \name.

\subsection{Details that Matter}
\label{sec:design_choices}
In this section, we will clarify details regarding choices and explanations that are previously skipped for coherence, including
the choice of the objective $\ell$,
the detailed setup of BN and 
the role of grouping in DBN.

\subsubsection{The Choice of The Objective $\ell$}
\label{sec:loss}
Here we will explain why we choose squared error $
    \ell_{SE} (z_1, z_2) = \| z_1 - z_2 \|_2^2
$ as our default setting instead of cosine similarity 
$
\ell_{cos}(z_1, z_2) = \frac{z_1^T z_2}{\|z_1 \|_2 \| z_2 \|_2},
$
which is more popular in self-supervised representation learning.

The main difference between maximizing cosine similarity $\ell_{cos}$ and minimizing squared error $\ell_{SE}$ is whether or not the vectors are normalized to unit $L_2$-norms, since $\ell_{SE}(z_1, z_2)= 2 - 2\ell_{cos}(z_1, z_2)$ when $\|z_1 \|_2=\|z_2\|_2 = 1$.

Such normalization may conflict with feature decorrelation. 
With squared error, for $i\in \{1,2\}, j\in \{1, \cdots, D\}$ where $z_1, z_2\in \mathcal{R}^D$, we have
\begin{align*}
    \frac{\partial \ell_{SE}}{\partial z_{i, j}} =  
    2 (z_{i, j} - z_{3-i, j}),
\end{align*}
 which involves only the $j$-th dimension itself.

However, with cosine similarity, for $i\in \{1,2\}, j\in \{1, \cdots, D\}$ where $z_1,  z_2\in \mathcal{R}^D$, we have
\begin{align*}
    \frac{\partial \ell_{cos}}{\partial z_{i, j}} = 
    \frac{z_{3-i, j}}{\|z_{3-i} \|_2} \cdot \left(
        \frac{1}{\| z_i \|_2} - \frac{z_{i, j}^2}{\| z_i\|_2^3} 
    \right)
    =\frac{z_{3-i, j}\sum_{k\neq j} z_{i, k}^2}{\|z_{3-i} \|_2\| z_i\|_2^3},
\end{align*}
which may depend heavily on other dimensions because of the normalization. This may introduce unnecessary interference in the projection space.

We suggest removing such normalization is favorable in feature decorrelation. Experimentally, \name fails in decorrelation with cosine similarity as the objective.

\subsubsection{The Detailed Setup of BN}
\label{sec:BN_affine_eps}
Here we will explain the detailed setup in using BN to avoid complete collapse. 
We include an empirical comparison of BN with different setups in Table \ref{tab:affine_eps}, where both the learnable affine transform and a non-negligible $\epsilon$ are detrimental to the utility of the learned representation.

The learnable affine transformation nullifies variance standardization simply because the variance of the output of BN scales linearly with the scaling parameter $\gamma$.

As for the $\epsilon$ designed originally for numerical stability, one should notice that for a given batch of inputs $X\in\mathcal{R}^{D\times B}$ with variance $\sigma^2 \in \mathcal{R}^D$, the variance of the $d$-th dimension of its output is in fact
\begin{align*}
    \hat{\sigma}_{d}^2 = \frac{\sigma_d^2}{\sigma_d^2 + \epsilon} 
    = 1 - \frac{\epsilon}{\sigma_d^2 + \epsilon},
\end{align*}
which is strictly monotonically increasing with $\sigma_d$ as long as $\epsilon > 0$ and therefore vanishing variance remains as a trivial, reachable solution.

Note that in Table \ref{tab:affine_eps}, we report in the comparison a setting with $\epsilon=0.1$, which is greater than typical choices and is used only as a proof of concept.

\begin{table}[!t]
    \centering
\resizebox{0.7\linewidth}{12mm}{
    \begin{tabular}{c c c}
        \toprule
        $\epsilon$ & learnable affine & acc.(\%) \\
        \midrule
        0  & No & 70.85\\
        0.1 & No & 34.47\\
        0   & Yes & 10.00 \\
        \bottomrule
    \end{tabular}
}
    \caption{A comparison of BN with different setups. Both the learnable affine transform and non-negligible $\epsilon$ are detrimental to eventual utility since they compromise variance standardization.}
    \label{tab:affine_eps}
\end{table}

\subsubsection{The Role of Grouping in DBN}
\label{sec:DBN_grouping}
The grouping strategy in DBN has two major benefits, one is for flexibility, and another is for efficiency. 

Recall that ZCA Whitening works only with the assumption that $\Sigma = \hat{X} \hat{X}^T \in \mathcal{R}^{D\times D}$ is full-rank. Otherwise, no linear transform on the features can result in a fully standardized covariance matrix, \ie an identity matrix $I$, since $I$ is full-rank. 

To have $\Sigma$ a matrix with rank $D$, $\hat{X} \in \mathcal{R}^{D\times B}$ must have a rank of at least $D$. Besides, since each row of $\hat{X}$ is zero-mean, we have its rank to be bounded by $B - 1$, which indicates that the minimum batch size $B$ allowed is at least $D+1$. This greatly limits the flexibility of ZCA Whitening as one has to either restrict the dimension of the feature space or scale the batch size linearly with it. With grouping, the batch size only has to scale with the group size $G$.

Another relatively minor benefit is the improved efficiency. Without grouping, a single pass of ZCA Whitening has a computational cost of $O(BD^2)$, with $B$ the batch size and $D$ the number of dimensions. In comparison, DBN with a group size of $G$ requires only a cost of $O(BDG)$.

\section{Evalution}
\label{sec:experiment}

\subsection{Experimental Setup}
\label{sec:setup}

\begin{table*}[!tbp]
\centering
   \begin{tabular}{c c c c c}
      \toprule
         & CIFAR-10 & CIFAR-100 & STL-10 & Tiny ImageNet \\
      \midrule
      SimCLR \cite{chen2020simple} & 86.96 & 55.86 & 85.50 & 42.65 \\
      BYOL \cite{richemond2020byol} & 86.65 & 59.33 & 85.59 & 42.75 \\
      SimSiam \cite{chen2020exploring} & 86.31 & 59.44 & \textbf{86.55} & 41.58 \\
      Barlow Twins \cite{zbontar2021barlow} & 89.02 & 62.84 & 85.43 & 45.33 \\ \hline
      DBN          & 86.32 & 56.49 & 82.36 & 40.37 \\
      \name & \textbf{89.50} & \textbf{62.95} & 86.02 & \textbf{45.96} \\
      \bottomrule
   \end{tabular}
   \caption{Top-1 accuracies($\%$) of DBN and \name in linear evaluation with 200-epoch pretraining. For completeness and reference, we include results of some representative methods from our reproduction. For a fair comparison, we use the same projector and augmentations as we describe in Section \ref{sec:setup} for all methods in the reproduction.}
   \label{tab:small}
\end{table*}

\begin{table}[!tbp]
    \centering
    \begin{tabular}{ccc}
        \toprule
                & CIFAR-10 & CIFAR-100\\
        
        \midrule
        SimCLR  & 75.05    & 50.42\\
        BYOL    & 78.63    & 51.44\\
        SimSiam & 78.77    & 50.71\\
   Barlow Twins & 79.54    & \textbf{57.22}\\
         \midrule
        DBN     & 78.60      & 52.95\\
        \name   & \textbf{80.62}    & 57.17\\
        \bottomrule
    \end{tabular}
    \caption{Top-1 accuracies($\%$) of DBN and \name in linear evaluation on CIFAR-10 and CIFAR-100 with 200-epoch pretraining on Tiny ImageNet. For completeness and reference, we include results of some representative methods from our reproduction. For a fair comparison, we use the same projector and augmentations as we describe in Section \ref{sec:setup} for all methods in the reproduction.}
    \label{tab:transfer}
\end{table}

\begin{itemize}[wide=0pt]
\setlength\itemsep{-0.25em}

    \item \textbf{Benchmarks.} We conduct extensive experiments on 5 popular benchmarks. 
    \textbf{CIFAR-10} and \textbf{CIFAR-100}~\cite{cifar} are two small-scale image datasets composed of 32$\times$32 small images with 10 and 100 classes, respectively.
    \textbf{STL-10}~\cite{coates2011analysis} and \textbf{Tiny ImageNet}~\cite{le2015tiny} are both medium-size datasets derived from the ImageNet dataset~\cite{russakovsky2015imagenet}. The STL-10 dataset is composed of 96 $\times$ 96 resolution images of 10 classes. For each class, STL-10 has 500 labeled training samples (5K labeled training samples in total) and 800 labeled samples for testing. An additional 100K unlabeled training images are sampled from a wider range of images than labeled ones. The Tiny ImageNet dataset has 200 classes and comprises 100K training data and 10K testing data, with 64 $\times$ 64 resolutions.
    \textbf{ImageNet ILSVRC-2012} is a popular large-scale image dataset of 1000 classes and 1.28M training images. It has 50K images for validation and 150K for testing.

   \item \textbf{Optimizer and learning rate}. Large-batch optimizers such as LARS~\cite{you2017large} are commonly used in self-supervised contrastive pre-training for visual representation learning~\cite{chen2020simple, grill2020bootstrap, Caron2020swav}. However, recent studies~\cite{Fetterman2020understanding, gaur2020training} indicate such adaptive gradient optimizers might regularize the network the same way batch norm does. To separate the inherent normalization properties of optimizers, we use SGD for pre-training. We set our base learning rate to be $0.02$ for experiments on small and medium-sized datasets and $0.06$ for large-scale datasets. We linearly scale the learning rate according to the batch size: $\frac{\text{base lr} \times \text{batch size}}{256}$~\cite{goyal2017accurate}. The learning rate is scheduled to a cosine decay rate and 5 warm-up epochs~\cite{loshchilov2016sgdr}. We keep the momentum parameter to be $0.9$. The weight decay rate is $0.001$ for small and medium-sized datasets and $\num{1e-4}$ for ImageNet.
   
   \item \textbf{Encoder-backbone.} ResNet-18 is adopted as the backbone of our encoder on small and medium datasets. For CIFAR-10 and CIFAR-100, we use the CIFAR variant of ResNet-18~\cite{he2016deep,chen2020exploring}, the first max-pooling layer of which is removed, and the kernel size of the first convolution layer is 3. For medium-size datasets STL-10 and Tiny ImageNet, only the max-pooling layer is disabled following~\cite{ermolov2020whitening,chen2020exploring}. We adopt a ResNet-50 as the encoder for large-scale ImageNet experiments. We remove the last fully connected layer in ResNet-18 and ResNet-50 models and treat the features after global average pooling as inputs to the projector.
   
   \item \textbf{Encoder-projector.} The projector is a 3-layer projection MLP. BN and ReLU activations are applied to all hidden layers of the projector. We set the hidden dimensions twice of the input dimension and keep the output dimension identical to the input dimension. Finally, we normalize the output using our \name layer. Unless otherwise specified, we set the group size of \name to be one-half of the batch size. 
   
   \item \textbf{Data augmentation.} We adopt several common data augmentations and compose them stochastically: (a) random scaling and cropping with a scaling factor chosen between $[0.2, 1.0]$; (b) random horizontal flipping with a probability of 0.5; (c) color distortion with a probability of 0.8; (d) color dropping (\ie, randomly convert images to grayscale with 20\% probability for each image); (e) random gaussian blur for medium and large-size datasets.
   
   \item \textbf{Training and Evaluation} We evaluate the quality of the pre-trained representations by training a supervised linear classifier on the frozen representations, following a common protocol. We perform unsupervised pre-training on the train set for 200 epochs. Then we freeze the features and train a supervised linear classifier, \ie, a fully-connected layer followed by a softmax layer, on the extracted features. Specifically, we train the linear layer on top of the global average pooling features of a ResNet for 100 epochs. To test the classifier, we use the center crop of the test set and computes accuracy according to predicted outputs. We train the classifier with a base learning rate of $30$, no weight decay, a momentum of $0.9$, and a batch size of $256$. Note that we only train the classifier over the labeled split of STL-10 since the majority of STL-10 training data is unlabeled. We report the validation accuracy for ImageNet. 
   
\end{itemize}

\subsection{Gains from Further Decorrelation}

In this section, to verify the gains from further decorrelation empirically, we have both DBN and \name evaluated on multiple benchmarks and have the results reported in Table \ref{tab:small}. 
Through further decorrelation, \name outperforms DBN on all $4$ benchmarks with performances competitive to the best of all evaluated methods, which supports the claim strongly.

In Table \ref{tab:transfer}, we also include a comparison of the generalizability of DBN and \name, by evaluating representations pretrained with Tiny ImageNet on CIFAR-10 and CIFAR-100. With further decorrelation, \name generalizes better than DBN in both cases, achieving performances competitive to the best one among all evaluated methods, just as in the prior setting.

\begin{table}[!tbp]
\resizebox{\linewidth}{9mm}{
    \centering
    \begin{tabular}{c c c c c c}
       \toprule
       dim.     & 64    & 128   & 256   & 512   & 1024\\
       \midrule
       DBN    & 77.17 & 82.15 & 82.42 & 82.91 & 84.39\\
       \name  & 82.92 & 83.19 & 84.54 & 86.02 & 87.22\\
       \bottomrule
    \end{tabular}
}
    \caption{
    Top-1 accuracies($\%$) in linear evaluation of DBN and \name on CIFAR-10 with different numbers of output dimension for a 3-layer MLP projector (\textbf{dim.}): In all cases, we use a group size of $32$ and a batch size of $256$; The hidden layers of the projector contain $1024$ neurons each.}
    \label{tab:ablation_dim}
\end{table}

In addition, we study the gains from further decorrelation varying the number of dimensions for the projection space. The results are in Table \ref{tab:ablation_dim}, from which one sees that further decorrelation yields further gains consistently, regardless of the specific choice of the projector's output dimensions.

Are the aforementioned gains generic varying the batch size? To answer this, we conduct an ablation study regarding the gains of feature decorrelation while using different batch sizes. The results are in Table \ref{tab:batch_size}, where \name offers top utilities in all cases, supporting strongly the generality of the gains from decorrelation.

\begin{table}[!tbp]
     \centering
      \begin{tabular}{c c c c c c}
      \toprule
      batch size & 32 & 64 & 128 & 256 & 512\\
      \midrule
      \name  & 88.25 & \textbf{89.17} & \textbf{89.31} & \textbf{88.82} & 87.92\\
      \hline
      Barlow Twins & 86.89 & 87.98 & 88.21 & 87.57 & 85.19\\
      BYOL & \textbf{88.37} & 88.44 & 87.64 & 85.72 & 82.63\\
      SimCLR & 85.42 & 87.41 & 87.40 & 87.70 & \textbf{87.98}\\
      SimSiam & 86.84 & 87.88 & 86.47 & 79.02 & 67.74\\
      \bottomrule
      \end{tabular}
      \caption{The top-1 accuracy($\%$) of \name and our own reproduction of Barlow Twins, BYOL, SimCLR and SimSiam at 200 epochs under linear evaluation on CIFAR-10. The training and evaluation configurations are the same. 2-layer MLP projectors with hidden dimension and output dimension to be 1024 and 512 are used for all experiments.}
      \label{tab:batch_size}
\end{table}

\subsection{Varying Decorrelation Strength}

\begin{table}[!tbp]
\centering
   \begin{tabular}{c c c c c}
   \toprule
    group size & 16 & 32 & 64 & 128\\
   \midrule
   kNN acc.     & 83.41& 85.93  & 87.05  & 87.59\\
   linear acc.   & 85.52  & 87.69 & 88.75 &  88.29\\
   \bottomrule
   \end{tabular}
   \caption{
        Accuracies($\%$) of \name on CIFAR-10 with different group size. \textbf{kNN acc.} denotes accuracy in kNN classification. \textbf{linear acc.} denotes accuracy in linear evaluation. The output dimension of the projector is $512$.
   }
   \label{tab:ablation_group_size}
\end{table}

Another way to verify the gains from further decorrelation is to vary the decorrelation strength of \name, which we achieve here by varying the group size $G$: The larger the group size $G$ is, the stronger the decorrelation strength will be. 

We report in Table \ref{tab:ablation_group_size} the results of such ablation study. We observe that the overall trend is consistent with our expectation in both kNN classification and linear evaluation: The utility improves with a stronger decorrelation strength.

\subsection{Feature Decorrelation on ImageNet}

\begin{table}[!tbp]
\centering
   \begin{tabular}{c c c}
      \toprule
      method & batch size & top-1 \\ 
      \midrule
      InstDisc \cite{wu2018unsupervised} & 256  & 58.5 \\
      LocalAgg \cite{zhuang2019local} & 128  & 58.8 \\
      MoCo \cite{he2020moco} & 256 & 60.6 \\
      SimCLR \cite{chen2020simple} & 256  & 61.9 \\
      CPC v2 \cite{oord2018representation} & 512  & 63.8 \\
      PCL v2 \cite{li2020prototypical} & 256  & 67.6 \\
       MoCo v2 \cite{chen2020improved} & 256  & 67.5 \\
       MoCHi \cite{MoCHi} & 512 & 68.0 \\
       PIC \cite{PIC} & 512 & 67.6 \\
       AdCo \cite{hu2020adco} & 256 & 68.6\\
       \hline
      Shuffled-DBN & 512 & 65.18 \\
      \bottomrule
   \end{tabular}
   \caption{Top-1 accuracies($\%$) in linear evaluation on ImageNet with the ResNet-50 backbone and 200 epochs of pretraining. The table are mostly inherited from \cite{hu2020adco}. Feature decorrelation in the concise framework acheives sub-optimal performance.}
   \label{tab:imgnet}
\end{table}

\begin{table}[!tbp]
\centering
\resizebox{\linewidth}{9mm}{
   \begin{tabular}{c c c c c c c}
   \toprule
      epoch  & 10  & 20 & 50    & 100  & 150  & 200  \\
   \midrule
   Top-1 & 49.12 & 54.32 & 57.43   & 59.04   & 62.24  & 65.18 \\
   Top-5 & 72.34 & 76.93 & 79.24   & 80.62   & 82.88  & 85.32\\
   \bottomrule
   \end{tabular}
}
   \caption{Top-1 and top-5 accuracies of \name in linear evaluation on ImageNet varying pre-training epochs.}
   \label{tab:checkpoint}

\end{table}

The accuracy in linear evaluation on ImageNet has become a de facto metric of visual features learned in self-supervised fashions. 
While the differences in both accessible computational resources and implementation details (\eg resources for hyperparameter tuning) are detrimental to the fairness of a direct comparison, we present in Table \ref{tab:imgnet} the evaluation on ImageNet and consider it as a nice addition to compare feature decorrelation with representative methods. 
For completeness, we also include in Table \ref{tab:checkpoint} the top-1 and top-5 accuracies of \name for checkpoints in the middle.

On ImageNet, although \name does not achieve state-of-the-art performances, it remains promising given that it achieves sub-optimal utility in a concise framework (\ie, with no predictor, no momentum encoder and no other special implementation detail).

\section{Conclusion}
In this work, we study the feature collapsing issues in self-supervised learning. Firstly, we verify the existence of complete collapse and address it by standardizing variance. 
Furthermore, we discover that an overlooked collapse pattern, namely dimensional collapse, is indeed reachable when learning representations in a self-supervised fashion. 
We connect dimensional collapse with strong correlations between axes and consider this connection a strong motivation for feature decorrelation (\ie standardizing the covariance matrix).

Through this work, we hope not only to present to our community the insights regarding the importance and the potential of feature decorrelation but also to facilitate future work that advances self-supervised learning by addressing design flaws instead of mostly trial and error.

{\small
\bibliographystyle{ieee_fullname}
\bibliography{egbib}
}

\end{document}